\documentclass[lettersize,journal]{IEEEtran}
\usepackage{amsmath,amsfonts}
\usepackage[colorlinks,
            linkcolor=red,
            anchorcolor=blue,
            citecolor=green
            ]{hyperref}
\usepackage{hyperref}
\usepackage{algorithm}
\usepackage{array}
\usepackage[caption=false,font=normalsize,labelfont=sf,textfont=sf]{subfig}
\usepackage{textcomp}
\usepackage{stfloats}
\usepackage{url}
\usepackage{verbatim}
\usepackage{graphicx}
\usepackage{cite}
\usepackage{algpseudocode}  
\usepackage[percent]{overpic}
\usepackage{caption}
\usepackage{enumitem}
\usepackage{tikz}
\usepackage{adjustbox}
\usepackage{cleveref}
\usepackage{multirow}
\begin{document}
\title{Part-aware Shape Generation with Latent 3D Diffusion of Neural Voxel Fields}

\author{Yuhang Huang, Shilong Zou, Xinwang Liu,~\IEEEmembership{Senior Member,~IEEE}, Kai Xu$^{*}$\thanks{$*$ Corresponding author (E-mail: kevin.kai.xu@gmail.com).},~\IEEEmembership{Senior Member,~IEEE}
\thanks{Y. Huang, S. Zou, X. Liu, and K. Xu are with the School of Computer, National University of Defense Technology, Changsha, 410073, China. 
}
}

\markboth{Journal of \LaTeX\ Class Files,~Vol.~14, No.~8, August~2021}%
{Shell \MakeLowercase{\textit{et al.}}: A Sample Article Using IEEEtran.cls for IEEE Journals}



\maketitle
\begin{abstract}
This paper introduces a novel latent 3D diffusion model for generating neural voxel fields with precise part-aware structures and high-quality textures. In comparison to existing methods, this approach incorporates two key designs to guarantee high-quality and accurate part-aware generation. On one hand, we introduce a latent 3D diffusion process for neural voxel fields, incorporating part-aware information into the diffusion process and allowing generation at significantly higher resolutions to capture rich textural and geometric details accurately. On the other hand, a part-aware shape decoder is introduced to integrate the part codes into the neural voxel fields, guiding accurate part decomposition and producing high-quality rendering results. Importantly, part-aware learning establishes structural relationships to generate texture information for similar regions, thereby facilitating high-quality rendering results. We evaluate our approach across eight different data classes through extensive experimentation and comparisons with state-of-the-art methods. The results demonstrate that our proposed method has superior generative capabilities in part-aware shape generation, outperforming existing state-of-the-art methods. Moreover, we have conducted image- and text-guided shape generation via the conditioned diffusion process, showcasing the advanced potential in multi-modal guided shape generation. 
\end{abstract}


\section{Introduction}
Recent generative models have emerged for generating 3D shapes with different representations \cite{get3d, wu2016learning, siddiqui2022texturify, henzler2019escaping, leng2018learning, guan2020fame, ye20213d, deng2021plausible, mao2021std, mo2019partnet, wang2020pie}, among which neural field representation \cite{pigan, eg3d, stylenerf, cips3d} receives increasing attention due to many advantages such as integrated texture and shape generation. Despite the fast and notable progress made along this line of research, what has been largely missed is part/structure awareness. Most existing generative models based on neural fields are part-oblivious. They tend to generate 3D shapes in a holistic manner, without comprehending their compositional parts explicitly.

Generating 3D shapes with part information facilitates several downstream processing/tasks, such as editing, mix-and-match modeling, and segmentation learning. 
SPAGHETTI~\cite{spaghetti} and Part-NeRF~\cite{partnerf} are two excellent part-aware generative models for 3D shape synthesis with neural field representation. Despite their strengths, there is still potential for enhancing the generative ability. SPAGHETTI relies on 3D supervision and mainly considers the geometry, ignoring the textured shape generation. On the other hand, Part-NeRF requires only 2D supervision but adopts a complex architecture with multiple neural radiance fields, leading to increased complexity in learning and subpar rendering outcomes. Moreover, both models perform object segmentation in an unsupervised manner, resulting in structurally insignificant information.

\begin{figure}
    \centering
    \includegraphics[width=\linewidth]{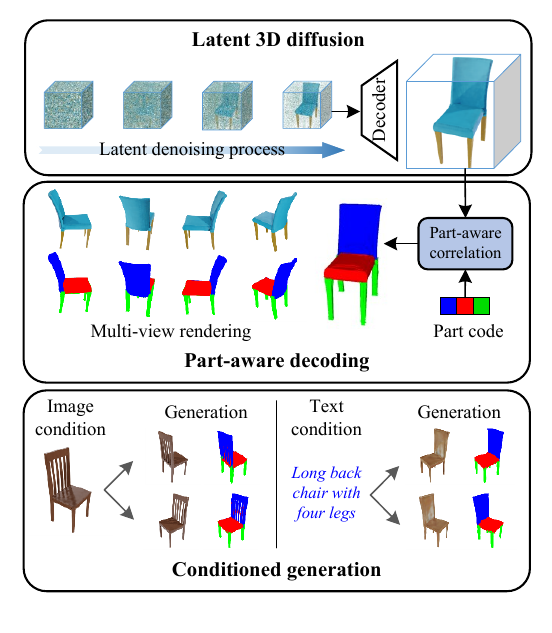}
    \caption{Our approach involves a latent 3D diffusion process and a part-aware decoding module. The latent 3D diffusion process facilitates the generation of high-resolution voxel fields, while the part-aware decoding module ensures precise part decomposition. Moreover, we admit multi-modal inputs for conditioned shape generation.}
    \label{fig:teaser}
\end{figure}

We propose a latent 3D diffusion model to generate neural voxel fields with accurate structural details in this paper. As shown in Fig~\ref{fig:teaser}, our network consists of two parts, a latent 3D diffusion model of neural voxel fields and a part-aware shape decoder. Specifically, the latent 3D diffusion model utilizes the advantage of 3D convolution and employs the 3D denoising process in a latent space with a much lower resolution. On the other hand, the part-aware shape decoder incorporates the part code with the neural voxel field to learn part-aware relationships, guiding the precise generation of part-aware shapes and synthesizing high-quality rendering results. Moreover, the part code is also incorporated into the diffusion model, which allows for learning the part-aware information in field generation.

Benefiting from the latent diffusion design, our model allows for shape generation with a much higher resolution than previous work (e.g., Matthias~\cite{diffrf} and \cite{karnewar2023holofusion}). In particular, our method achieves $96^3$ resolution which is sufficient for representing rich geometric details of most common-seen objects. Furthermore, we integrate cross- and self-attention mechanisms into the part-aware decoder to establish robust dependencies between the part code and the voxel field. This integration aids in modeling the inherent relationships among various parts, significantly enhancing the precision of part segmentation. More importantly, learning part-aware information enhances the correlations between different regions, facilitating geometry reconstruction and texture modeling in similar parts.

\begin{figure*}[tb]
    \centering
    \begin{overpic}
        [width=1.0\textwidth]{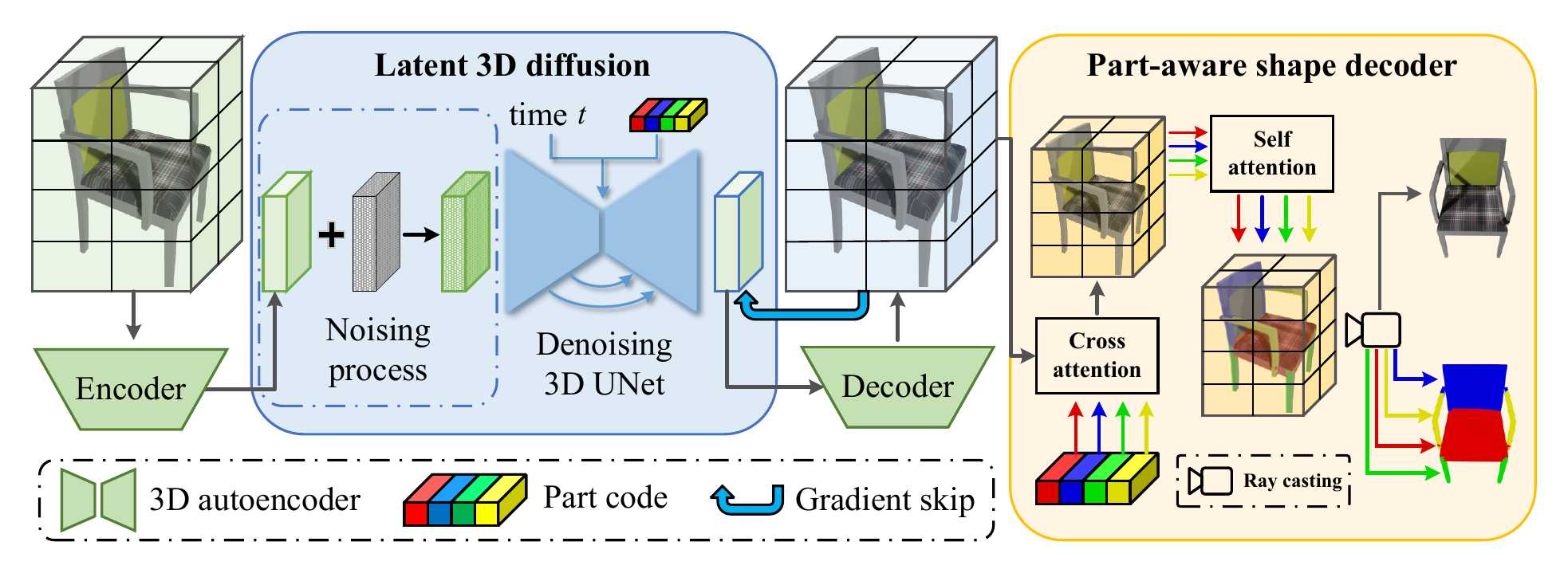}
        \put(17.3, 26.7){\small $\mathbf{L}_{0}$}
        \put(23, 26.7){\small $\boldsymbol{\epsilon}$}
        \put(28, 26.7){\small $\mathbf{L}_{t}$}
         \put(45.7, 26.7){\small $\mathbf{L}_{\boldsymbol{\phi}}$}
    \end{overpic} 
    \caption{The proposed method comprises two main modules: a latent 3D diffusion model and a part-aware shape decoder. The latent 3D diffusion model facilitates the high-resolution generation of neural voxel fields, while the part-aware shape decoder enables the generation of parts-aware results for rendering. The two modules are trained jointly.}
    \label{fig1}
\end{figure*}
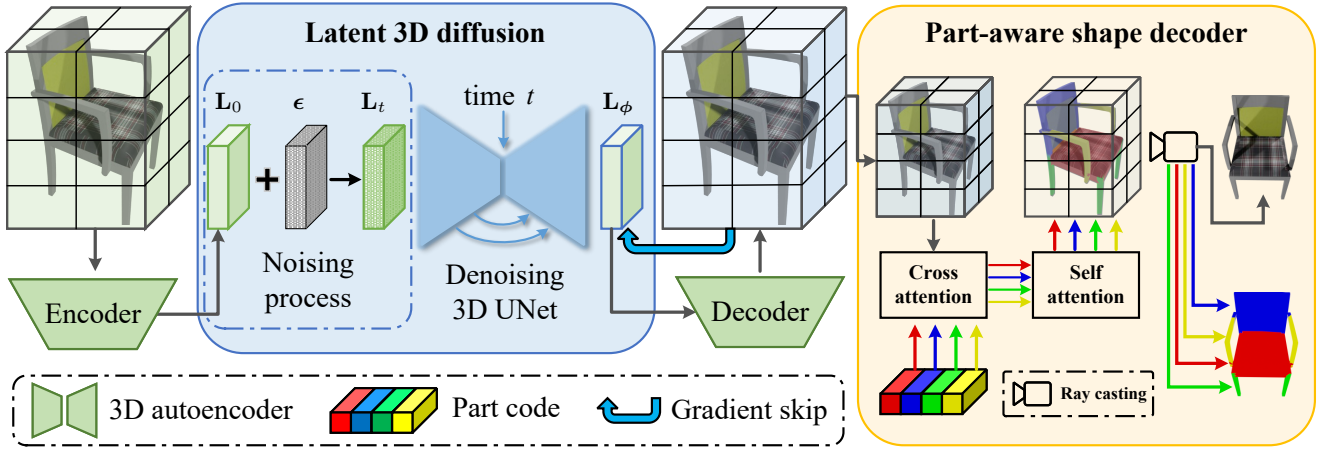 

Our method adopts full volume learning part-learning and decoder-based part decomposition. This avoids the design of per-part neural fields which makes the decomposition and rendering overly complicated. Our architecture makes rendering rather simplistic and allows for high-quality rendering and hence accurate training.

The diffusion model admits multi-modal prompts for shape generation. For example, we have implemented image-conditioned diffusion enabling single-image guided generation. Concretely, a convolutional encoder is employed to extract features from the reference image. Subsequently, these obtained features are combined with the multi-scale features of the denoising UNet. As a result, the image information can effectively guide the shape generation. Moreover, we perform text-conditioned diffusion for text-guided shape generation. We utilize the general language model CLIP \cite{radford2021clip} to extract embedding for the text input, and perform cross-attention between the text embedding and the multi-scale features of the denoising UNet.

To make a fair comparison, we have implemented a supervised version of Part-NeRF. Similar to acquiring color, we use MLP layers to implicitly learn the part class probability of 3D position, obtaining the semantic part map via volume rendering. We conduct extensive experiments and compare to the state-of-the-art methods on eight classes of common-seen objects. The results show that our method achieves the best metrics (FID, MMD, COV) on all classes of data and higher-quality part-aware generation.

\section{Related Work}

\noindent\textbf{Diffusion Models for Shape Generation}. 
Recent developments in diffusion models \cite{ddpm} have enabled the synthesis of 3D shapes through a variety of representations, including point cloud \cite{achlioptas2018learning, sharma2020parsenet, liu2022efficient}, SDF \cite{chou2023diffusion}, mesh \cite{get3d}, and neural field \cite{pigan, epigraf, voxgraf, devries2021unconstrained, liao2020towards}. 
\cite{hu2024neural} proposes to utilize diffusion models to generate a compact wavelet representation with a pair of coarse and detail coefficient volumes. 
3DShape2VecSet \cite{zhang20233dshape2vecset} introduces a novel representation for neural fields designed for generative diffusion models, which can encode 3D meshes or point clouds as neural fields. DiffFacto \cite{nakayama2023difffacto} learns the distribution of point clouds with part-level control including part-level sampling, mixing and interpolation. EXIM \cite{liu2023exim} use the text prompt as guidance, generating corresponding truncated signed distance field with diffusion models.
However, most of above methods must adopt 3D input data as supervision, which is not available easily. Differently, this paper focus on generating neural voxel fields with differentiable rendering, enabling multi-view image as supervision.

\textbf{Diffusion Models for Generating Neural Fields}. 
Recently, the diffusion models \cite{sohl2015deep, ddpm} have outperformed GANs in various 2D tasks \cite{diffusionbeatgan, cao2024animediffusion}, which has inspired several diffusion-based methods to generate neural fields. DreamFusion \cite{dreamfusion} utilizes the 2D pretrained diffusion model as the distillation model to guide the generation of neural fields for the first time. Following this idea, a series of researches \cite{qian2023magic123, long2023wonder3d, zhou2023sparsefusion} can edit and generate neural fields via text prompt. However, they need to retrain the model for each generation, leading to a time-consuming generation process. Differently, another line of work directly applies the diffusion model to synthesize the neural fields without 2D pretrained diffusion models. DiffRF \cite{diffrf} applies the diffusion model to generate neural voxel fields, however, the 3D convolutional network takes up too much memory cost, limiting the generated resolution. SSD-NeRF \cite{ssdnerf} proposes a single-stage training paradigm to generate tri-plain fields with 2D convolutional network. In this paper, we propose a 3D latent diffusion model to directly generate neural voxel fields, enabling high-resolution and high-quality generation.


\textbf{Part-aware Shape Generation}. 
Previous methods \cite{spgan, anise, pqnet, spaghetti, dualsdf, koo2023salad, zhu2020adacoseg} usually generate part-aware shapes with point cloud representation, however, they typically rely on 3D supervision. Recent PartNeRF \cite{partnerf} proposes to represent the different parts of a shape with different neural radiance fields, however, the divided parts are meaningless and multiple neural fields require much high computational resources. Differently, we carefully design the part relation module and supervise the model with 2D part mask, achieving precise and high-quality part-aware shape generation.

\section{Proposed Method}

In this section, we introduce a latent 3D diffusion model to generate high-resolution part-aware neural voxel fields. As illustrated in Fig.~\ref{fig1}, our approach consists of a latent 3D diffusion model and a part-aware shape decoder. Specifically, the latent 3D diffusion model compresses the object fields into the latent space with a much smaller spatial resolution, and then adopts the 3D UNet to conduct the diffusion process. This implementation facilitates the high resolution of the voxel field, allowing the high-quality rendering results. Moreover, the part code is fed into the diffusion model to incorporate part-aware information into field generation. On the other hand, the part-aware shape decoder integrates the part codes into the voxel field, establishing relationships between different parts of the voxel field, and produces RGB images and semantic part maps through ray casting. 
\begin{figure*}
    \centering
    \includegraphics[width=\textwidth]{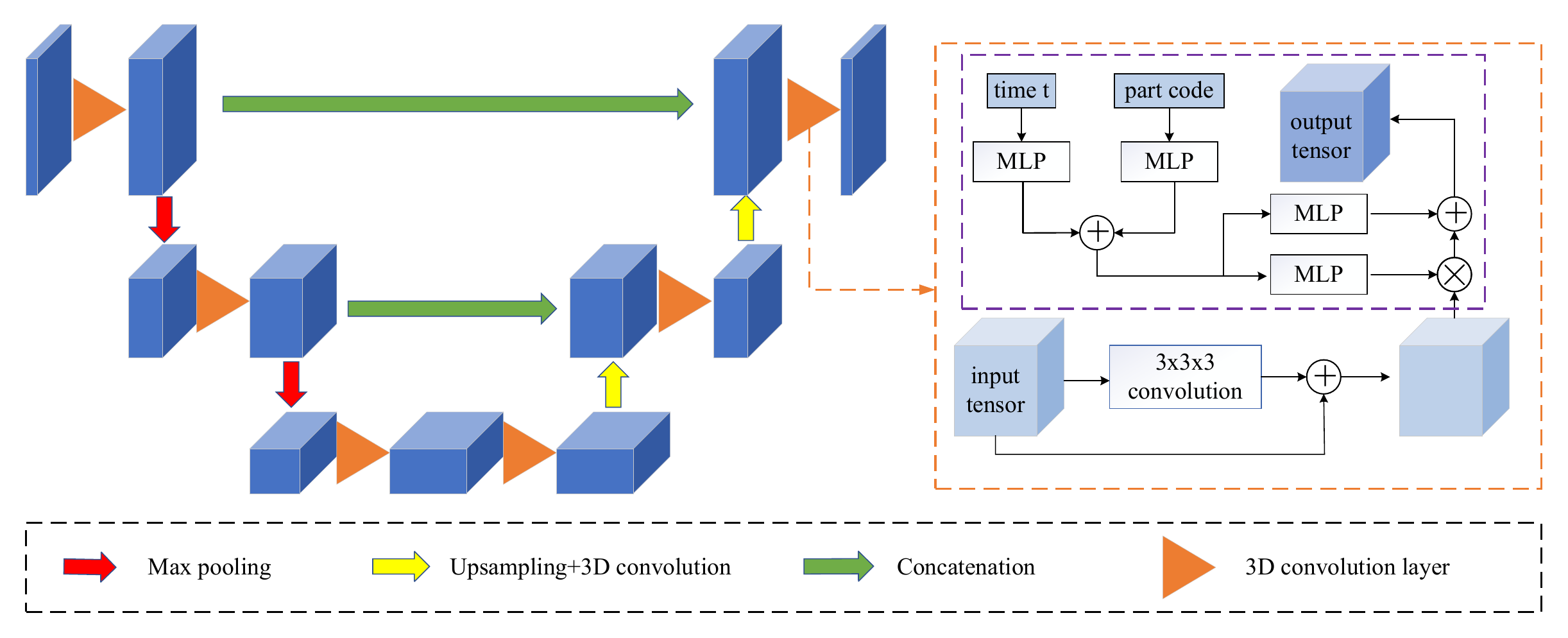}
    \caption{Detailed architecture of the 3D UNet. The 3D autoencoder does not have the modules in the dashed purple rectangular box.}
    \label{fig:arch}
\end{figure*}

\subsection{Preliminary}
\subsubsection{Neural Voxel Fields}
Neural radiance fields (NeRFs) \cite{nerf} represent 3D objects implicitly and map the 3D position $\boldsymbol{x}$ and view direction $\boldsymbol{d}$ to the density $\sigma$ and color emission $\boldsymbol{c}$, obtaining the novel view image via volume rendering. Based on NeRFs, DVGO \cite{dvgo} learns neural voxel fields to represent the object's density and color. Specifically, there are a density voxel field and a feature voxel field, which are represented by $\mathbf{V}^{\sigma}$ and $\mathbf{V}^{\boldsymbol{c}}$. We can acquire the density and color feature of position $\boldsymbol{x}$ by querying from $\mathbf{V}^{\sigma}$ and $\mathbf{V}^{\boldsymbol{c}}$ directly, and the color feature and view direction are feed into multilayer perceptron (MLP) layers to get the color emission $\boldsymbol{c}$.
\begin{equation}
\begin{aligned}
    \text{query}(\mathbf{V}, &\boldsymbol{x}): (\mathbb{R}^{C\times X\times Y\times Z}, \mathbb{R}^{3})\rightarrow \mathbb{R}^{C},\\
    &\sigma = \text{query}(\mathbf{V}^{\sigma}, \boldsymbol{x}), \\
    \boldsymbol{c} = &\text{MLP}(\text{query}(\mathbf{V}^{\boldsymbol{c}}, \boldsymbol{x}); \boldsymbol{d}).
\end{aligned}
\end{equation}
Here, query($\mathbf{V}, \boldsymbol{x}$) denotes the function querying the value of position $\boldsymbol{x}$ from the voxel field $\mathbf{V}$, $C$ is the dimension of the modality, and $(X, Y, Z)$ is the spatial size of $\mathbf{V}$.
With the ray casting, the RGB color of a given ray $\boldsymbol{r}$ can be rendered by the following equation:
\begin{equation}
\begin{aligned}
    \hat{\boldsymbol{c}}(\boldsymbol{r})& = \sum_{i=1}^{M}
    {(\prod_{j=1}^{i-1}{(1 - \alpha_{j})})\alpha_{i}\boldsymbol{c}_{i}},\\
    &\alpha_{i} = 1 - \exp(-\sigma_{i}\delta_{i}),
\end{aligned}
\label{eq2}
\end{equation}
where $M$ is the number of sampled points along the ray, $\alpha_{i}$ is the probability of the termination at point $i$, and $\delta_{i}$ is the distance between the sampled point and its adjacent point.
In this way, the neural voxel field can be optimized by minimizing the mean square error of the rendered RGB $\hat{\boldsymbol{c}}(\boldsymbol{r})$ and the ground truth color $\boldsymbol{g}_{\boldsymbol{c}}(\boldsymbol{r})$. After the optimization, we can represent the object with $\mathbf{V}^{\sigma}$ and $\mathbf{V}^{\boldsymbol{c}}$. To simplify the formulation, we concatenate $\mathbf{V}^{\sigma}$ and $\mathbf{V}^{\boldsymbol{c}}$ along the first dimension, resulting in $\mathbf{V}^{\boldsymbol{f}}$ as the representation of the neural voxel field.

\subsubsection{Decoupled Diffusion Models}
Diffusion models have shown great potential in generative tasks, however, they usually suffer from prolonged inference time. Recent decoupled diffusion models \cite{ddm} propose to decouple the diffusion process to speed up the sampling process. The decoupled forward diffusion process is controlled by the analytic data attenuation and the standard Wiener process:
\begin{equation}
    q(\mathbf{V}^{\boldsymbol{f}}_{t}|\mathbf{V}^{\boldsymbol{f}}_{0}) = \mathcal{N}(\mathbf{V}^{\boldsymbol{f}}_{0}+\int_{0}^{t}{\boldsymbol{a}_{t}\mathrm{d}t}, t\mathbf{I}),
    \label{eq3}
\end{equation}
where $t\in[0, 1]$ is the time step, $\mathbf{V}^{\boldsymbol{f}}_{0}=\mathbf{V}^{\boldsymbol{f}}$ is the original voxel field and $\mathbf{V}^{\boldsymbol{f}}_{t}$ is the noisy field at time $t$. $\boldsymbol{a}_{t}$ is the analytic attenuation function describing the process of the original voxel field to zero field, and $\mathbf{I}$ is the identity matrix.
According to the derivation of \cite{ddm}, the corresponding reversed process is written as:
\begin{equation}
\begin{aligned}
  q(\mathbf{V}^{\boldsymbol{f}}_{t-\Delta t}|\mathbf{V}^{\boldsymbol{f}}_{t}, \mathbf{V}^{\boldsymbol{f}}_{0}) = \mathcal{N}(\mathbf{V}^{\boldsymbol{f}}_{t}-&\int_{t-\Delta t}^{t}{\boldsymbol{a}_{t}\mathrm{d}t}-\frac{\Delta t}{\sqrt{t}}\boldsymbol{\epsilon}, \\
  &\frac{\Delta t(t-\Delta t)}{t}\mathbf{I}),
\end{aligned}
\end{equation}
where $\Delta t$ is the time interval, $\boldsymbol{\epsilon}\sim\mathcal{N}(\mathbf{0}, \mathbf{I})$ is the standard normal noise. Benefiting from the analytic attenuation function, we can solve the reversed process with arbitrary time intervals via analytic integration, improving the sampling efficiency greatly.
In practice, we choose the particular constant form of $\boldsymbol{a}_{t}$: $\boldsymbol{a}_{t}=-\mathbf{V}^{\boldsymbol{f}}_{0}$.
In this way, the model is supervised by $\boldsymbol{\epsilon}$ and $\mathbf{V}^{\boldsymbol{f}}_{0}$ simultaneously, and the training objective is formulated by: 
\begin{equation}
    \min\limits_{\boldsymbol{\theta}} \mathbb{E}_{q(\mathbf{V}^{\boldsymbol{f}}_{0})} \mathbb{E}_{q(\boldsymbol{\epsilon})} [\Vert \boldsymbol{a}_{\boldsymbol{\theta}}+\mathbf{V}^{\boldsymbol{f}}_{0}\Vert^{2} + \Vert \boldsymbol{\epsilon}_{\boldsymbol{\theta}}-\boldsymbol{\epsilon}\Vert^{2}],
\end{equation}
where $\boldsymbol{\theta}$, $\boldsymbol{a}_{\boldsymbol{\theta}}$ and $\boldsymbol{\epsilon}_{\boldsymbol{\theta}}$ are the parameter and two outputs of the diffusion model respectively.

\subsection{Latent 3D Diffusion Model}
Inspired by the preliminary work, we introduce a latent 3D diffusion model for the generation of high-resolution neural voxel fields. Our model offers two key advantages in comparison to previous diffusion-based methods. Unlike conventional 2D diffusion models \cite{ddpm, ddim, diffusionbeatgan, ldm}, we utilize 3D convolution to directly denoise the voxel field in 3D space, which enhances our modeling capabilities in terms of 3D geometry and consistency. 
On the other hand, due to the substantial computational demands of 3D convolution, training a 3D diffusion model with high-resolution voxel fields proves challenging, thereby constraining the generative quality. To address this limitation, we shift the denoising process to a latent space with reduced resolution, effectively eliminating constraints on voxel field resolution.

As shown in Fig.~\ref{fig1}, the proposed latent 3D diffusion model comprises a 3D autoencoder and a 3D denoising UNet. The 3D autoencoder has an encoder for compressing the voxel field $\mathbf{V}^{\boldsymbol{f}}_{0}$ to the latent code $\mathbf{L}_{0}$ and a decoder for recovering the original voxel field from $\mathbf{L}_{0}$, i.e. $\mathbf{L}_{0}=\mathcal{E}(\mathbf{V}^{\boldsymbol{f}}_{0}), \hat{\mathbf{V}}^{\boldsymbol{f}}_{0}=\mathcal{D}(\mathbf{L}_{0})$. Here, $\mathcal{E}$ and $\mathcal{D}$ represent the encoder and decoder of the 3D autoencoder; $\hat{\mathbf{V}}^{\boldsymbol{f}}_{0}$ is the reconstructed voxel field. In practice, the resolution of $\mathbf{L}_{0}$ is 64$\times$ lower than $\mathbf{V}^{\boldsymbol{f}}_{0}$.   
After obtaining the latent code, we adopt a 3D UNet architecture to conduct the diffusion process in the latent space. Our 3D UNet not only takes the noisy latent code and time as input, but also introduces the part code to incorporate structural information. In practice, the part code is a learnable embedding and can be represented by $\boldsymbol{z}\in\mathbb{R}^{K\times D}$, where $K$ is the number of parts, $D$ is the dimension of the embedding. Similar to Eq.~\ref{eq3}, we sample the noisy code $\mathbf{L}_{t}$ by adding the normal noise $\boldsymbol{\epsilon}$ to $\mathbf{L}_{0}$. Then, the noisy code is fed into the 3D UNet with the time $t$ and part code $z$ together, outputting the estimations of $\mathbf{L}_{0}$ and $\boldsymbol{\epsilon}$. This process is formulated by:
\begin{equation}
\begin{aligned}
    &\mathbf{L_{t}} = (1 - t)\mathbf{L}_{0} + \sqrt{t}\boldsymbol{\epsilon},\\
    &\mathbf{L}_{\boldsymbol{\phi}}, \boldsymbol{\epsilon}_{\boldsymbol{\phi}} = \text{UNet}_{\boldsymbol{\phi}}(\mathbf{L}_{t}, t, z),
\end{aligned}
\end{equation}
where $\boldsymbol{\phi}$ is the parameter of 3D UNet, $\mathbf{L}_{\boldsymbol{\phi}}$ and $\boldsymbol{\epsilon}_{\boldsymbol{\phi}}$ are the predictions of $\mathbf{L}_{0}$ and $\boldsymbol{\epsilon}$.
Note that we use the special form of the attenuation function: $\boldsymbol{a}_{t}=-\mathbf{L}_{0}$.

For the conditioned generation, we utilize the classifier-free \cite{ho2022classifier} manner to synthesize shapes corresponding to the conditioned input. Specifically, the conditioned input is fed into a feature extractor, resulting in a feature embedding. We adopt the feature embedding as an additional input of the 3D UNet, thus we obtain the $\mathbf{L}_{\boldsymbol{\phi}}$ and $\boldsymbol{\epsilon}_{\boldsymbol{\phi}}$ by: $\mathbf{L}_{\boldsymbol{\phi}}, \boldsymbol{\epsilon}_{\boldsymbol{\phi}} = \text{UNet}_{\boldsymbol{\phi}}(\mathbf{L}_{t}, t,\mathbf{e})$, where $\mathbf{e}$ is the feature embedding of the conditioned input.


\textbf{3D UNet vs. 2D UNet}. Though the latent code $\mathbf{L}_{t}$ is a 4D tensor, it is possible to apply the 2D convolutional operator to this type of data. In practice, we can reshape $\mathbf{L}_{t}$ into a 3D tensor, i.e. $\text{reshape}(\mathbb{R}^{C\times X\times Y\times Z})\rightarrow\mathbb{R}^{CX\times Y\times Z}$, thus adopting 2D UNet to model the denoising process. However, we found that the loss curve did not converge and it failed to obtain satisfactory results (details can be found in Sec.~\ref{sec4.3}). We assume that the 2D UNet architecture hardly learns the 3D consistency of the voxel field, resulting in a perturbed result. In contrast, the 3D UNet architecture provides reasonable performance, demonstrating the necessity of adopting 3D UNet for denoising the neural voxel field.

\begin{table*}[th]
  \caption{Quantitative comparisons to state-of-the-art methods on generative rendering performances. Note the KID metrics are multiplied by 1000.
  }
  \label{tab1}
  \setlength{\tabcolsep}{5pt}
  \renewcommand{\arraystretch}{1.2}
  \centering
\resizebox{1\textwidth}{!}{
\begin{tabular}{lcccccccccccccccc}
\hline
\multirow{2}{*}{Method} & \multicolumn{2}{c}{Chair}         & \multicolumn{2}{c}{Table}         & \multicolumn{2}{c}{Airplane}      & \multicolumn{2}{c}{Car}           & \multicolumn{2}{c}{Guitar}        & \multicolumn{2}{c}{Lamp}          & \multicolumn{2}{c}{Laptop}        & \multicolumn{2}{c}{Pistol}        \\
                        & FID$\downarrow$ & KID$\downarrow$ & FID$\downarrow$ & KID$\downarrow$ & FID$\downarrow$ & KID$\downarrow$ & FID$\downarrow$ & KID$\downarrow$ & FID$\downarrow$ & KID$\downarrow$ & FID$\downarrow$ & KID$\downarrow$ & FID$\downarrow$ & KID$\downarrow$ & FID$\downarrow$ & KID$\downarrow$ \\ \hline
Part-NeRF               & 69.44           & 14.29           & 75.78           & 26.91           & 83.84           & 42.10           & 133.2           & 77.73           &   123.16       &   79.78      &   106.44     &    71.09      &   237.02        &  203.34     &    137.4      &   90.57        \\
EG3D                &  104.24     &    36.71      &   116.88       &    76.47      &   108.82      &   64.77     &  162.79    &   128.52    &     111.83         &     67.85        &  131.52        &   70.24         &   232.65       &     198.93       &  126.33        &     80.00       \\
DiffRF                  & 43.92           & 7.25            & 56.28           & 22.61           & 53.33           & \textbf{19.85}  & 88.33           & \textbf{26.65}  &    80.55     &    49.25       &     99.77         &    43.48        &   198.32        &     107.64        &    65.83         &      45.19       \\
Ours                    & \textbf{27.73}  & \textbf{5.12}   & \textbf{44.25}  & \textbf{21.76}  & \textbf{35.94}  & 20.53           & \textbf{61.39}  & 30.15           &   \textbf{62.71}    &        \textbf{29.98}  &   \textbf{75.40}      &   \textbf{16.54}       &  \textbf{141.88}  &  \textbf{80.54}    &    \textbf{61.33}      &   \textbf{33.35}     \\ \hline
\end{tabular}
}
\end{table*}

\subsection{Part-aware Shape Decoder}
After the denoising process of the latent 3D diffusion model, we can have a neural voxel field containing the density and color features. However, it can not provide the structural information and rendering results. Therefore, we propose a part-aware shape decoder to generate part-aware details and rendering results. 
As shown in Fig.~\ref{fig1}, to model the part-aware information, the shape decoder integrates the part code $z$ into the voxel field and utilizes the attention mechanism to build the relations between different parts of the voxel field.
To incorporate the part code into the neural field, we first project the reconstructed voxel field $\hat{\mathbf{V}}^{\boldsymbol{f}}_{0}$ to have the same dimension as $\boldsymbol{z}$, and then flatten $\hat{\mathbf{V}}^{\boldsymbol{f}}_{0}$ to be a 2D tensor like $\mathbb{R}^{XYZ\times D}$. Thus, we can utilize the multi-head cross-attention \cite{attention} to build a strong dependency between the part code and the voxel field. Moreover, to divide the voxel field accurately, we employ a self-attention to model the long-range relations between the different parts of the voxel field. After the cross-attention and self-attention, we reshape the tensor back to be a 4D voxel field. We formulate this process by the following equation:
\begin{equation}
    \overline{\mathbf{V}}^{\boldsymbol{f}}_{0} = \text{MHSA}(\text{MHCA}(\boldsymbol{z}, \hat{\mathbf{V}}^{\boldsymbol{f}}_{0})), \overline{\mathbf{V}}^{\boldsymbol{f}}_{0}\in\mathbb{R}^{D\times X\times Y\times Z},
\end{equation}
where MHSA and MHCA represent the multi-head self-attention and cross-attention.

Incorporated with the part code, the refined voxel field $\overline{\mathbf{V}}^{\boldsymbol{f}}_{0}$ contains both the color and structural information. In this way, we can utilize volume rendering to generate images and part maps of arbitrary views. Given a 3D position $\boldsymbol{x}$, we can query the point feature of the voxel field, and acquire the color emission and part class probability with MLP layers.
\begin{equation}
    \begin{aligned}
        \boldsymbol{c} &= \text{MLP}(\text{query}(\overline{\mathbf{V}}^{\boldsymbol{f}}_{0}, \boldsymbol{x}); \boldsymbol{d}; \boldsymbol{z}), \\
        &\boldsymbol{p} = \text{MLP}(\text{query}(\overline{\mathbf{V}}^{\boldsymbol{f}}_{0}, \boldsymbol{x}); \boldsymbol{z}),
    \end{aligned}
\end{equation}
where $\boldsymbol{p}\in\mathbb{R}^{K}$ is the part class probability. In contrast to Eq.~\ref{eq3}, we incorporate the part code into the MLP layers collectively to encode the structural information into individual points, thereby enhancing the generative intricacies. Subsequently, we derive the RGB image and semantic part map via ray casting following Eq.~\ref{eq2}.

\begin{figure*}[p]
    \centering
    \includegraphics[width=0.96\textwidth]{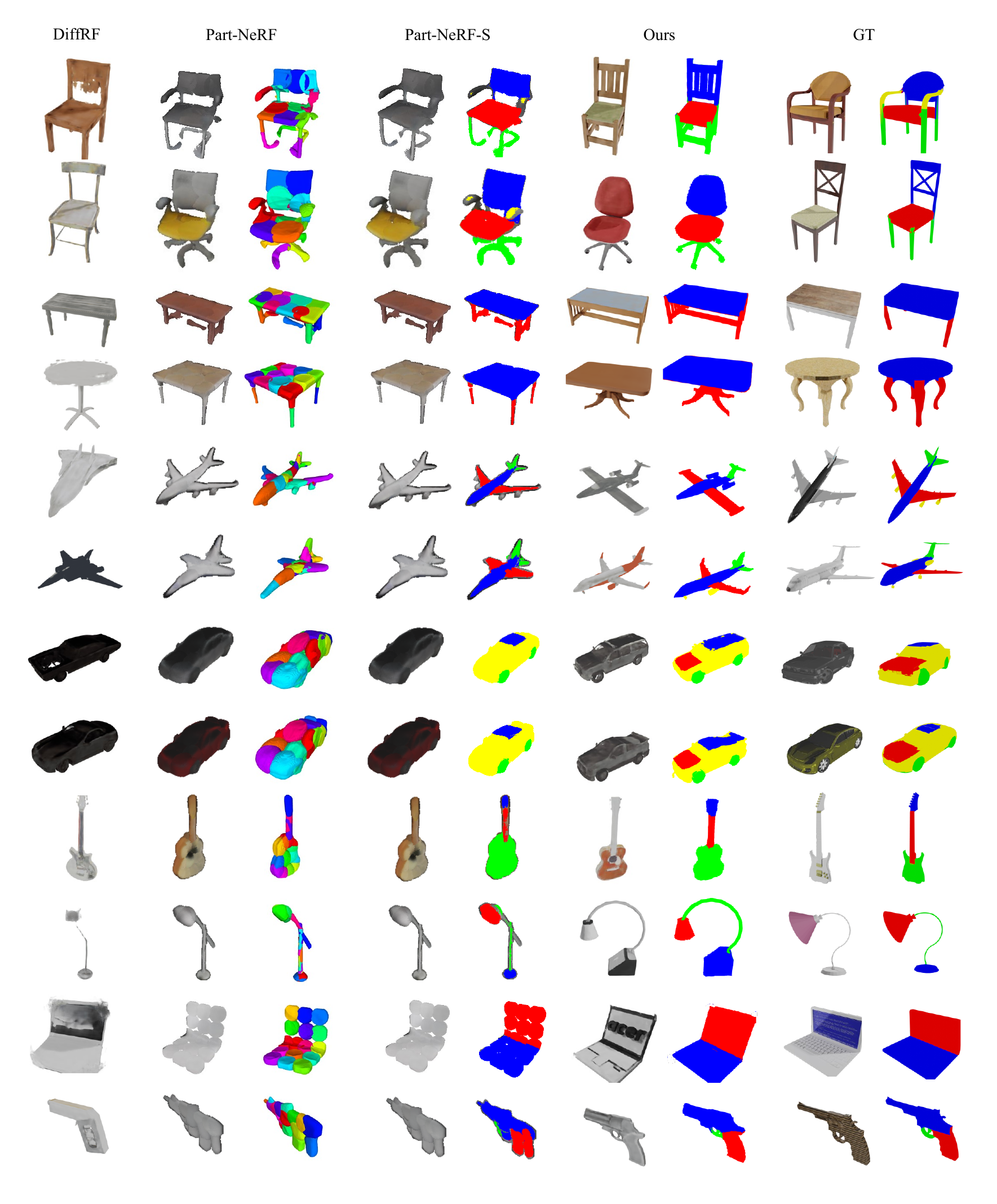}
    \caption{Qualitative comparison against state of the art. Part-NeRF-S denotes adding the 2D part supervision to the original Part-NeRF.}
    \label{fig3}
\end{figure*}

\subsection{End-to-end Training with Gradient Skip}
Several methods \cite{functa, gaudi} in the literature have combined the diffusion model with the neural field decoder, typically training these two components independently. However, independent training neglects the consideration of noise in the prior term $\hat{\mathbf{V}}^{\boldsymbol{f}}_{0}$ during the training of the neural field decoder, leading to biased learning of the decoder. In contrast, we propose an end-to-end training approach for the latent 3D diffusion model and the part-aware shape decoder, thereby avoiding noisy generation.

\textbf{Gradient Skip}. The diffusion model is trained on the latent space while the part-aware shape decoder is applied to the voxel field directly. Besides, there is the decoder $\mathcal{D}$ of the 3D autoencoder between the two modules connecting the latent space and the voxel field space. Obviously, training the two modules jointly needs the backward propagation of the gradient of $\mathcal{D}$. However, this implementation incurs significant computational costs and hinders the effective feedback of gradients. To mitigate this problem, we propose to skip the gradient of $\mathcal{D}$,  allowing the gradient of the part-aware shape decoder to directly propagate back to the latent space. Concretely, according to the Chain Rule, we have:
\begin{equation}
    \begin{aligned}
        \nabla_{\boldsymbol{\phi}}{\mathcal{L}_{rend}} = \frac{\partial\mathcal{L}_{rend}}{\partial\hat{\mathbf{V}}^{\boldsymbol{f}}_{0}}
	\frac{\partial\hat{\mathbf{V}}^{\boldsymbol{f}}_{0}}{\partial\mathbf{L}_{\boldsymbol{\phi}}}
	\frac{\partial\mathbf{L}_{\boldsymbol{\phi}}}{\partial\boldsymbol{\phi}}, 
    \end{aligned}
\end{equation}
where $\mathcal{L}_{rend}$ is the rendering loss.
We ignore the term $\partial\hat{\mathbf{V}}^{\boldsymbol{f}}_{0}/\partial\mathbf{L}_{\boldsymbol{\phi}}$ to skip the gradient of the decoder in the 3D autoencoder, resulting in an efficient formulation of gradient calculation:
\begin{equation}
        \nabla_{\boldsymbol{\phi}}{\mathcal{L}_{rend}} =  \frac{\partial\mathcal{L}_{rend}}{\partial\hat{\mathbf{V}}^{\boldsymbol{f}}_{0}}
	\frac{\partial\mathbf{L}_{\boldsymbol{\phi}}}{\partial\boldsymbol{\phi}}.
\end{equation}\label{eq:10}
Since the shape of $\mathbf{L}_{\boldsymbol{\phi}}$ is different from $\hat{\mathbf{V}}^{\boldsymbol{f}}_{0}$, we use trilinear interpolation to align their shapes for calculating the above equation.

The main training objective is the summation of the diffusion loss and the rendering loss.
\begin{equation}
\begin{aligned}
    &\mathcal{L} = \mathcal{L}_{diff} + \lambda_{t}\mathcal{L}_{rend},\\
    \mathcal{L}_{diff} &= \Vert{\mathbf{L}_{\phi}-\mathbf{L}_{0}}\Vert^{2} + \Vert{\boldsymbol{\epsilon}_{\phi}-\boldsymbol{\epsilon}}\Vert^{2},\\
    \mathcal{L}_{rend} = \sum_{\boldsymbol{r}\in\mathcal{R}}&\Vert\hat{\boldsymbol{c}}(\boldsymbol{r})-\boldsymbol{g}_{\boldsymbol{c}}(r)\Vert^{2}+\text{CE}(\hat{\boldsymbol{p}}(\boldsymbol{r}), \boldsymbol{g}_{\boldsymbol{p}}(r)),
\end{aligned}
\end{equation}
where $\hat{\boldsymbol{c}}(\boldsymbol{r})$ and $\hat{\boldsymbol{p}}(\boldsymbol{r})$ are the rendered RGB image and part map from ray $\boldsymbol{r}$, and $\boldsymbol{g}_{\boldsymbol{c}}(r)$ and $\boldsymbol{g}_{\boldsymbol{p}}(r)$ are the corresponding ground truths; CE means the cross-entropy loss function. Since the predicted voxel field of the diffusion model is reasonable only with the time $t$ close to zero, we attach a time-dependent weight $\lambda_{t}$ to the rendering loss: $\lambda_{t} = -\log(t)$.

Except for the main training objective, we adopt two additional regularization loss functions. First, we conduct the total variation on the neural voxel fields to restrain the noise generation:
\begin{equation}
\footnotesize
    \mathcal{L}_{tv} = \sum_{dim\in[D]}{\sqrt{\Delta_{x}^{2}(\hat{\mathbf{V}}^{\boldsymbol{f}}_{0}, dim)+\Delta_{y}^{2}(\hat{\mathbf{V}}^{\boldsymbol{f}}_{0}, dim)+\Delta_{z}^{2}(\hat{\mathbf{V}}^{\boldsymbol{f}}_{0}, dim)}},
\end{equation}
where $\Delta_{x}^{2}(\hat{\mathbf{V}}^{\boldsymbol{f}}_{0}, dim)$ denotes the squared difference between the value of $dim$-th channel in voxel: $v:=(i;j;k)$ and the $dim$-th value in voxel $(i+1;j;k)$, which can be extended to $\Delta_{y}^{2}(\hat{\mathbf{V}}^{\boldsymbol{f}}_{0}, dim)$ and $\Delta_{z}^{2}(\hat{\mathbf{V}}^{\boldsymbol{f}}_{0}, dim)$. Moreover, we constrain the value of part code $\boldsymbol{z}$ via its L2 norm $\Vert\boldsymbol{z}\Vert_{2}$ attached with the weight 0.0001.

\subsection{Detailed Architectures}
We adopt the 3D UNet architecture following \cite{cciccek20163d} and 
add the additional time and part code input branches. More specifically, the time and part code are projected to be a scale weight and a bias weight, and the two weights are used to transform the intermediate variable of the 3D convolution layer. The 3D autoencoder is also a 3D UNet architecture without additional branches. The detailed architecture of the 3D UNet is depicted in Fig.~\ref{fig:arch}. 


\section{Experiments}
\subsection{Experimental Setup}
\subsubsection{Dataset} We collect eight categories of 3D objects from ShapeNet \cite{shapenet}, including Chair, Table, Airplane, Car, Lamp, Guitar, Laptop, and Pistol. 
For each category, we random sample 500 objects, with a split of 450 for training and 50 for testing purposes. Utilizing BlenderProc \cite{blenderproc}, we render 100 views of each object at a resolution of 800$\times$800 resolution. With the posed images, we use the public code base \cite{dvgo} to generate neural voxel fields. For text-guided shape generation, we get the text labels of Chair and Table from \cite{chen2019text2shape}. For ground truth part labels, we use the shape segmentation method \cite{feng2019meshnet} to obtain the labeled shapes, and then use BlenderProc to render 2D segmentation maps.

\begin{table*}[tb]
  \caption{Quantitative comparisons to state-of-the-art methods on MMD and COV using Chamfer distance. Note the metrics are multiplied by 100.
  }
  \label{tab:2}
  \setlength{\tabcolsep}{3pt}
  \renewcommand{\arraystretch}{1.2}
  \centering
  \resizebox{1\textwidth}{!}{
    \begin{tabular}{lcccccccccccccccc}
\hline
\multirow{2}{*}{Method} & \multicolumn{2}{c}{Chair}       & \multicolumn{2}{c}{Table}       & \multicolumn{2}{c}{Airplane}    & \multicolumn{2}{c}{Car}         & \multicolumn{2}{c}{Guitar}      & \multicolumn{2}{c}{Lamp}        & \multicolumn{2}{c}{Laptop}      & \multicolumn{2}{c}{Pistol}      \\
                        & MMD$\downarrow$ & COV$\uparrow$ & MMD$\downarrow$ & COV$\uparrow$ & MMD$\downarrow$ & COV$\uparrow$ & MMD$\downarrow$ & COV$\uparrow$ & MMD$\downarrow$ & COV$\uparrow$ & MMD$\downarrow$ & COV$\uparrow$ & MMD$\downarrow$ & COV$\uparrow$ & MMD$\downarrow$ & COV$\uparrow$ \\ \hline
Part-NeRF               &   1.17        &   60.4        &  1.26           &  56.4         &  0.48         &  39.6        &  0.64          &  23.8         &  0.26          &  31.7         &   1.57         &  {59.4}      &  0.86          &  31.7         &   0.34          &   48.5        \\
EG3D                &     2.79        &    52.4       &     2.08    &    49.9     &     1.45        &    40.9       &    0.81         &   24.0        &     0.97        &    30.0       &    1.76         &   55.6        &     1.18        &    31.1       &    0.45         &   48.9        \\
DiffRF                  &    1.10        &    58.9      &   1.28         &  55.7        &    0.35       &   60.3   &    0.60       &  25.9    &      0.30       &     35.6      &   1.61          &    56.3       &     0.92        &   33.5         &   0.38           &     50.0       \\
Ours                    & \textbf{1.00}       & \textbf{61.5}     & \textbf{1.13}       & \textbf{59.9}     & \textbf{0.18}       &  \textbf{82.3}    & 0.46       &  \textbf{42.1}        & \textbf{0.14}        &  \textbf{45.1}      &   \textbf{1.48}       &     \textbf{59.7}    &  \textbf{0.79}        &   \textbf{38.8}     & \textbf{0.31}         &  \textbf{50.5}       \\ \hline
\end{tabular}
}
\end{table*}

\begin{table*}[tb]
  \caption{Quantitative comparisons to state-of-the-art methods on MMD and COV using Earth Mover's distance. Note the metrics are multiplied by 100.
  }
  \label{tab:3}
  \setlength{\tabcolsep}{3pt}
  \renewcommand{\arraystretch}{1.2}
  \centering
  \resizebox{1\textwidth}{!}{
    \begin{tabular}{lcccccccccccccccc}
\hline
\multirow{2}{*}{Method} & \multicolumn{2}{c}{Chair}       & \multicolumn{2}{c}{Table}       & \multicolumn{2}{c}{Airplane}    & \multicolumn{2}{c}{Car}         & \multicolumn{2}{c}{Guitar}      & \multicolumn{2}{c}{Lamp}        & \multicolumn{2}{c}{Laptop}      & \multicolumn{2}{c}{Pistol}      \\
                        & MMD$\downarrow$ & COV$\uparrow$ & MMD$\downarrow$ & COV$\uparrow$ & MMD$\downarrow$ & COV$\uparrow$ & MMD$\downarrow$ & COV$\uparrow$ & MMD$\downarrow$ & COV$\uparrow$ & MMD$\downarrow$ & COV$\uparrow$ & MMD$\downarrow$ & COV$\uparrow$ & MMD$\downarrow$ & COV$\uparrow$ \\ \hline
Part-NeRF               &   13.2       &   52.5        &   13.1          &  52.5         &  10.3           &  27.7         &  10.0          &   23.7       &   7.78          &   26.7        &  17.4           & 54.5          &  11.4          &  31.8         &   9.96          &   32.7        \\
EG3D                &    20.2         &  38.5         &     18.5        &   47.6         &    9.39         &   42.0        &  14.2           &   22.1        &    10.5         &    30.3       &  21.9           &   43.9        &    16.0         &    31.9       &  9.98           &   34.1        \\
DiffRF                  &     13.5        &    50.8       &  13.4          &   54.3       &      9.95       &   30.3     &     9.41       & 28.8     &      7.01       &      30.5      &    17.8         &    53.9      &     13.4        &   32.0       &     9.52       &     36.1      \\
Ours                    & \textbf{11.9}       & \textbf{54.0}     & \textbf{12.1}       & \textbf{57.4}     & \textbf{7.00}       &  \textbf{51.6}        & \textbf{8.66}       &   \textbf{37.0}       &  \textbf{6.35}         &  \textbf{37.4}       &  \textbf{16.0}       &    \textbf{55.3}    &   \textbf{11.2}       &  \textbf{36.9}      &  \textbf{9.29}      &  \textbf{36.9}      \\ \hline
\end{tabular}
}
\end{table*}

\begin{figure}[tb]
    \centering
    \includegraphics[width=\linewidth]{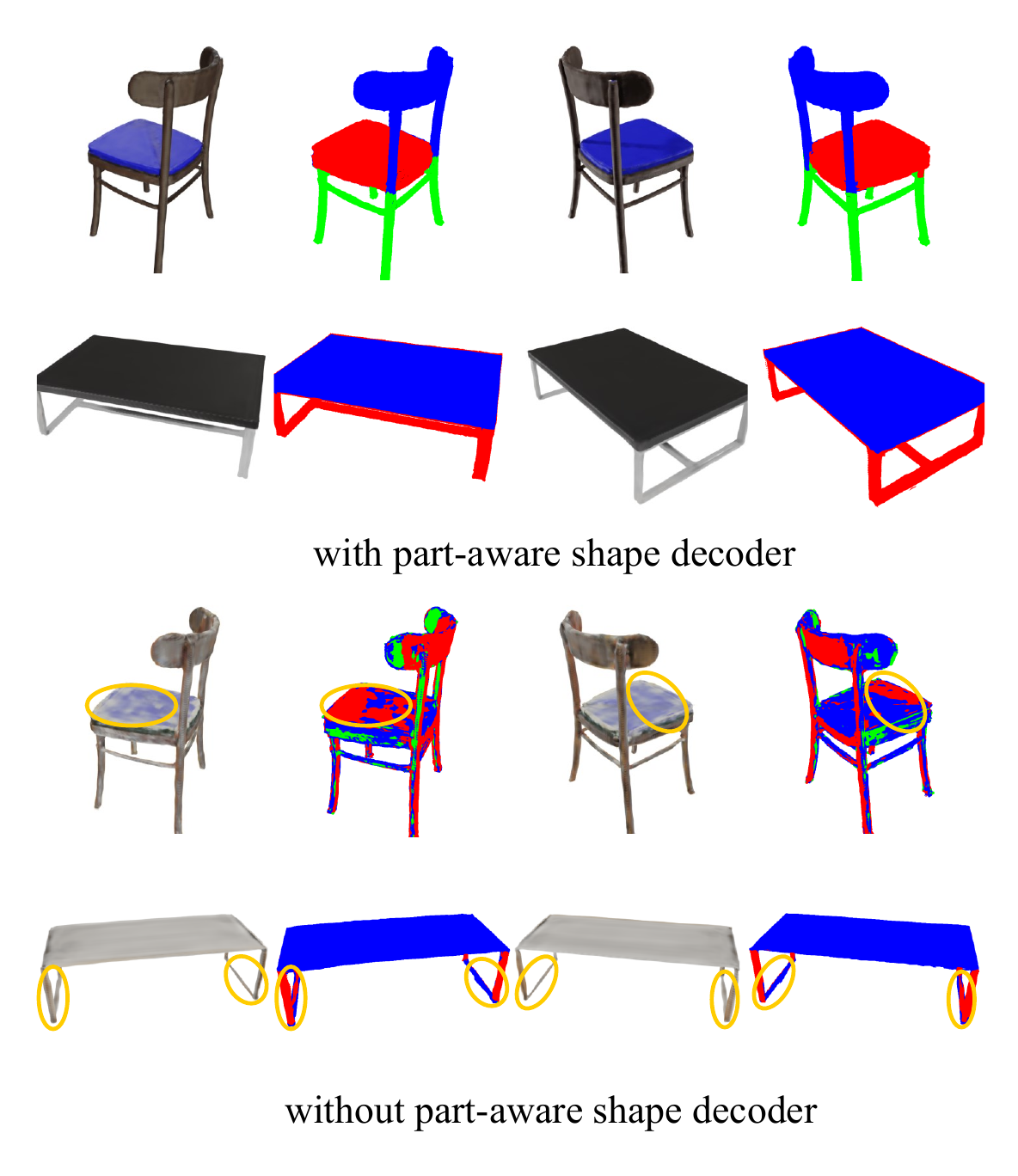}
    \caption{Visual improvements of the part-aware shape decoder. Without the part-aware shape decoder, there are inconsistent rendering results in the yellow circles, which shows that part-aware information benefits texture learning.}
    \label{fig4-2}
\end{figure}

\begin{figure}[!h]
    \centering
    \includegraphics[width=\linewidth]{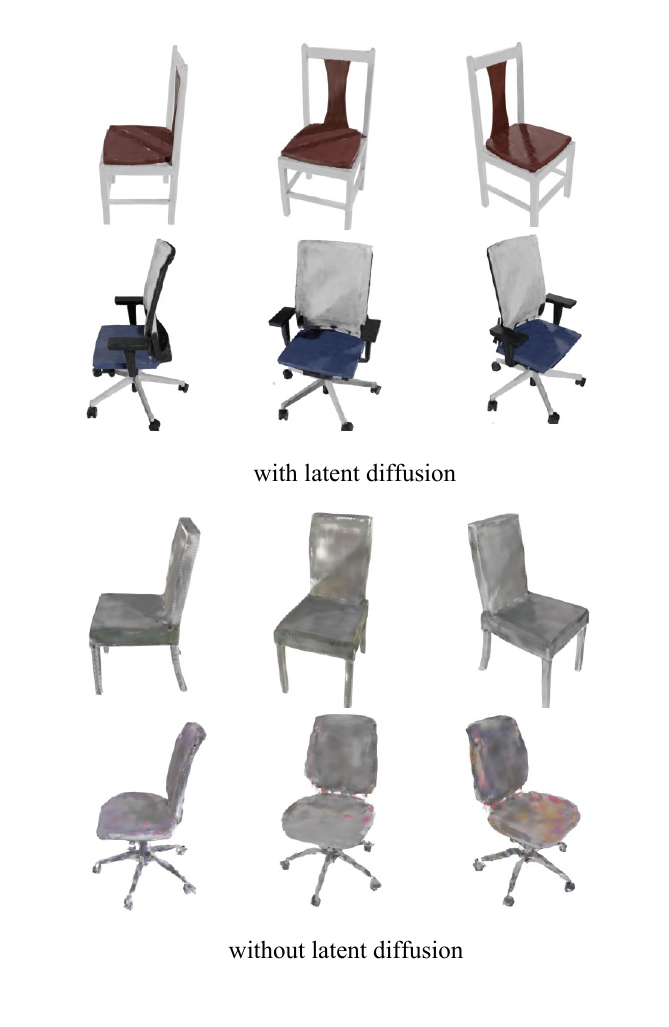}
    \caption{Visual improvements of the latent diffusion.}
    \label{fig4-1}
\end{figure}

\subsubsection{Implementation Details} The resolution of the voxel field is 96$^3$, and the resolution of the latent code is 64$\times$ lower than the voxel field. We train the model for 100,000 iterations and we do not train the part-aware shape decoder for the first 20,000 iterations to warm up the latent 3D diffusion model. We set $K=4$ and $D=128$ for the part code. For each iteration, we sample 8192 rays of 5 random views to calculate the rendering loss. For the architecture of 3D UNet, we set the base channel to 96 and the channel multiplier is [1, 2, 4], resulting in 144M parameters of the whole model. We train the model in a single RTX 3090 GPU with batch size 2, taking up around 15,000M GPU memory cost. The 3D autoencoder is pretrained on the same NeRF dataset as the whole model, and we apply a Kullback-Leibler Divergence loss on the latent codes to regularize its distribution close to the normal distribution following \cite{rombach2022high}.

\subsubsection{Conditioned Generation} We performed two types of modality for conditioned generation: image and text. For image condition, we use Swin-S \cite{swintrans} as the image feature extractor to process the single image, and then concatenate the obtained features with the multi-scale features of the denoising UNet. For text condition, we utilize CLIP \cite{radford2021clip} to extract the text embedding and correlate it with the multi-scale features of the denoising UNet via the cross-attention mechanism.

\subsubsection{Evaluation} We report the Frechet Inception Distance (FID) and Kernel Inception Distance (KID) to evaluate the quality of rendered images. We evaluate the two metrics at a resolution of 800$\times$800. 
In practice, we generate 200 objects and 10 views for each object to calculate metrics.
We also report the Coverage (COV) and the Minimum Matching Distance (MMD) \cite{AchlioptasDMG18} using Chamfer distance and Earth Mover's distance. MMD measures how likely it is that a generated shape looks like a test shape. COV measures how many shape variations are covered by the generated shapes.

\begin{figure*}[tb]
    \centering
    \includegraphics[width=\textwidth]{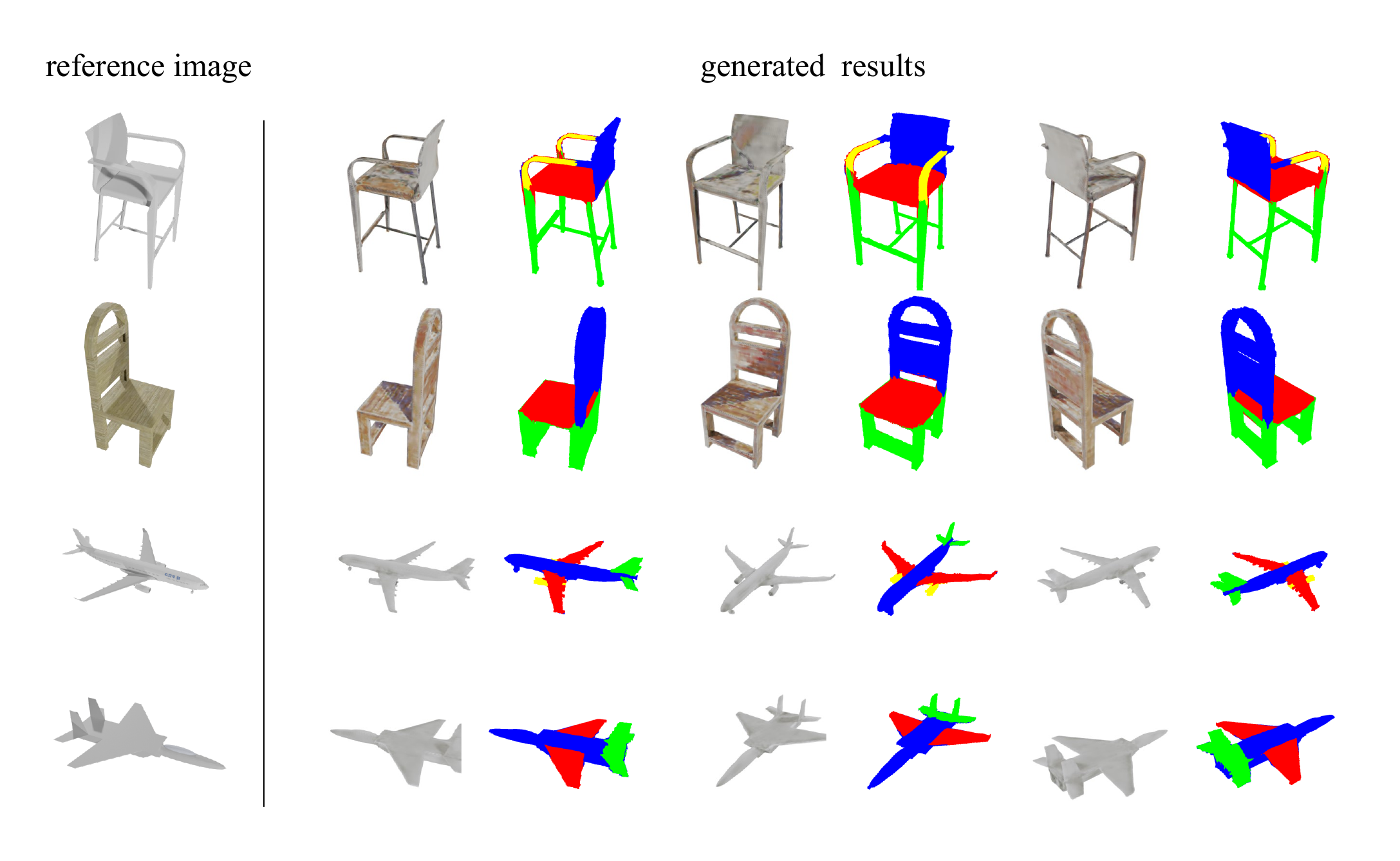}
    \caption{Single image to shape generation. As depicted in the figure, the generated shape has a similar geometry as the reference image but with different textures, which indicates the model learns more geometry information from the reference image.}
    \label{fig7}
\end{figure*}

\begin{figure*}[tb]
    \centering
    \includegraphics[width=\textwidth]{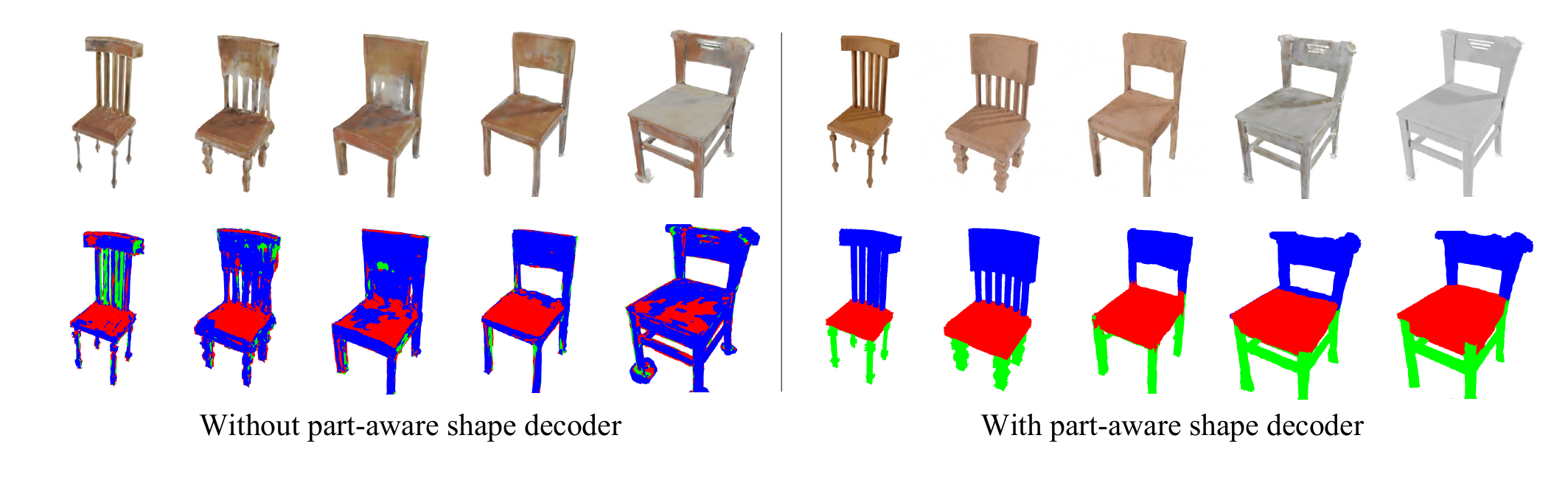}
    \caption{Visualization of shape interpolation. The generated intermediate shapes also have an accurate part-aware structure.}
    \label{fig5}
\end{figure*}

\subsection{Shape Generation}
\textbf{Quantitative comparison against state of the art}. In Tab.~\ref{tab1}, Tab.~\ref{tab:2} and Tab.~\ref{tab:3}, We have conducted a quantitative comparison between our proposed method and state-of-the-art approaches. Tab.~\ref{tab1} reports the metrics of generated image quality, while Tab.~\ref{tab:2} and Tab.~\ref{tab:3} report the geometry metrics calculated by Chamfer distance and Earth Mover's distance. 
EG3D \cite{eg3d} presents a GAN-based framework that introduces a sophisticated hybrid explicit-implicit network architecture for synthesizing the triplane representation. Despite its lightweight nature, the triplane representation struggles to encompass intricate details, resulting in unsatisfied rendering quality and geometry.
Part-NeRF \cite{partnerf} employs multiple neural radiance fields to represent a single object, but the complex and redundant architecture increases the learning difficulty, ultimately yielding lower-quality results. 
DiffRF \cite{diffrf} also utilizes the diffusion model in 3D space; however, it directly applies the 3D UNet to the voxel field. This implementation imposes significant computational demands, constraining the resolution of the voxel fields and leading to sub-optimal performances. Leveraging the 3D latent diffusion design enables the generation of 3D shapes with enhanced geometric details, leading to the attainment of the highest MMD and COV metrics across all classes. Additionally, the part-aware shape decoder facilitates the establishment of relationships between distinct parts, thereby improving the rendering of different components and resulting in the best FID metric across all classes.

\textbf{Qualitative comparison against state of the art}. As depicted in Fig.~\ref{fig3}, we plot the visual generations of various state-of-the-art methods. Leveraging the latent 3D diffusion model, we can generate high-resolution voxel fields, resulting in superior rendering quality. Similar to ours, Part-NeRF also produces part-aware rendering results. However, it employs an unsupervised approach that leads to the synthesis of meaningless structures. To ensure a fair comparison, we employ semantic part maps to supervise Part-NeRF, denoting this as Part-NeRF-S. As shown in Fig.~\ref{fig3}, benefiting from the design of latent diffusion and part-aware shape decoder, our approach still outperforms Part-NeRF-S in both texture and part quality.  Both quantitative and qualitative results indicate that the proposed method can generate accurate part-aware shapes and render high-quality images.


\begin{table}[tb]
\caption{Ablation study of the latent diffusion on Chair/Airplane. `LD' denotes `latent diffusion'. Here we use the Chamfer distance to calculate MMD and COV.}
        \label{tab:4}
        \centering
        \setlength{\tabcolsep}{2pt}
        \renewcommand{\arraystretch}{1.3}
        \resizebox{1\linewidth}{!}
        {
        \begin{tabular}{lcccccc}
\hline
\multirow{2}{*}{Method} & \multicolumn{3}{c}{Chair}                            & \multicolumn{3}{c}{Airplane}                         \\
                        & FID $\downarrow$ & MMD $\downarrow$ & COV $\uparrow$ & FID $\downarrow$ & MMD $\downarrow$ & COV $\uparrow$ \\ \hline
with LD                 & 27.73            &          1.00   &        61.5    & 35.94            &      0.18        &     82.3       \\
without LD              & 48.18            &    1.40          &  55.1         & 56.19            &    0.40          &    71.0       \\ \hline
\end{tabular}
        }
\end{table}

\begin{figure*}[tb]
    \centering
    \includegraphics[width=1\textwidth]{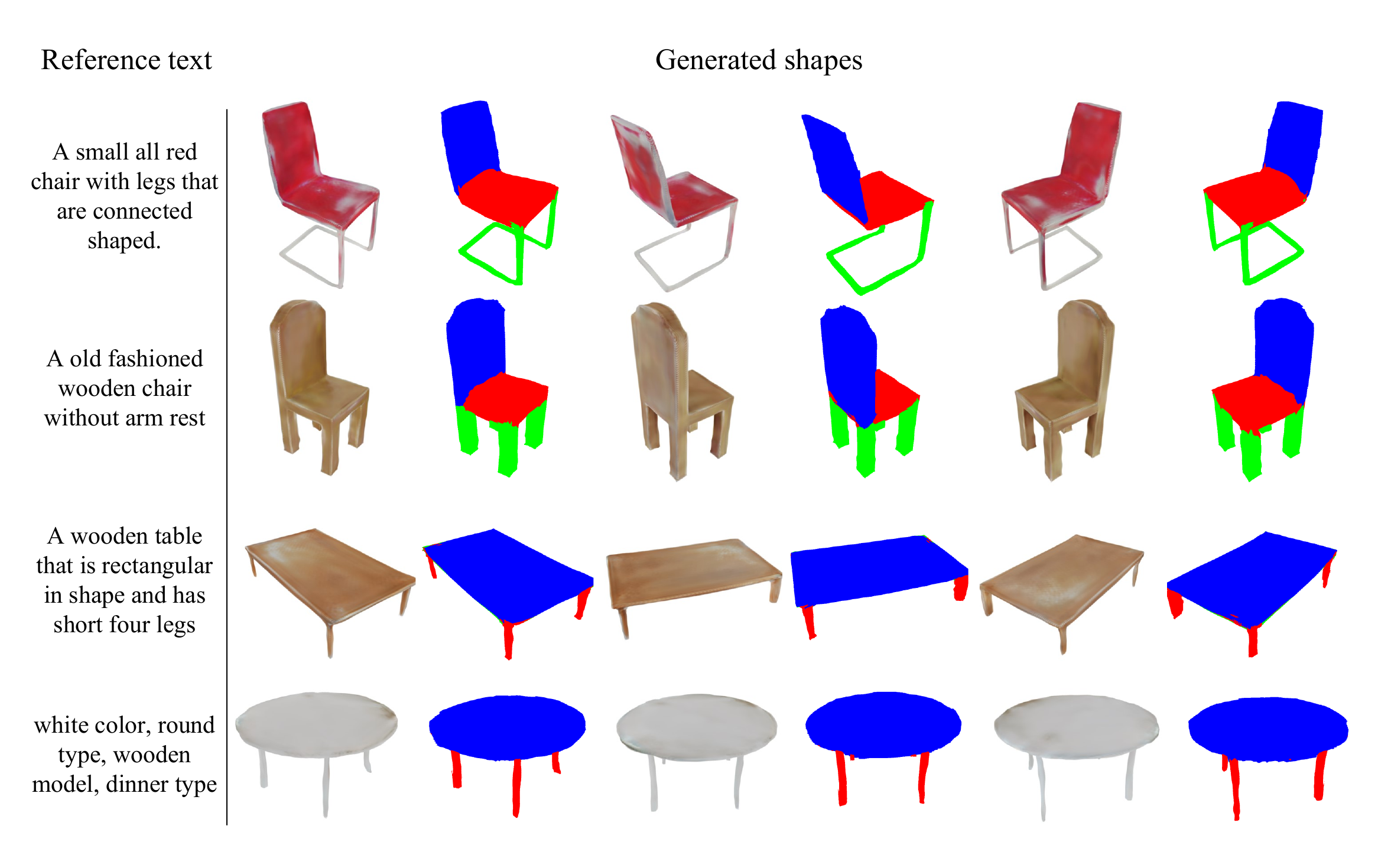}
    \caption{Text-guided shape generation. Given the specific descriptions, we can generate the corresponding shapes.}
    \label{fig:text}
\end{figure*}

\begin{figure}[t]
        \includegraphics[width=\linewidth]{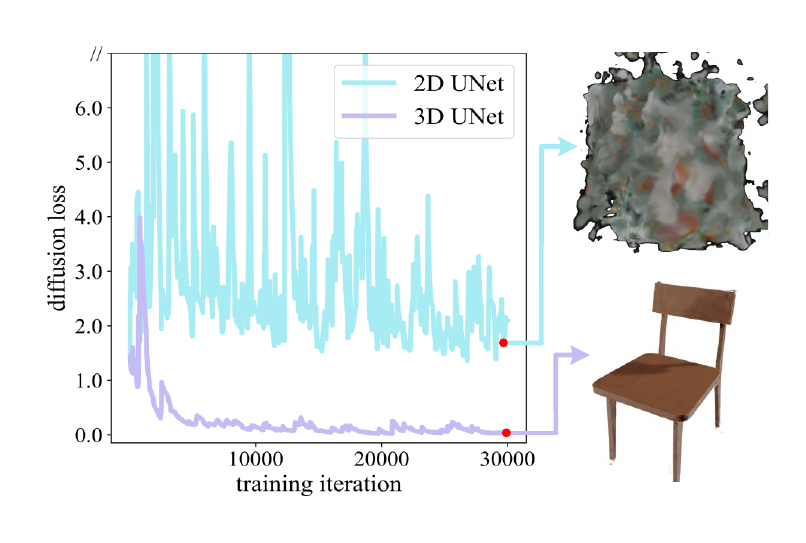}
    \caption{Comparison between different denoising networks. The left sub-figure depicts the training curves of 2D UNet and 3D UNet, while the right sub-figure shows the visual results of their generations.}
    \label{fig2}
\end{figure}

\subsection{Ablation Studies}\label{sec4.3}
We ablate different components of the proposed method to show their importance.


\begin{table}[t]
\caption{Ablation of part aware shape decoder on Chair/Airplane. `PASD' denotes `part-aware shape decoder'. Here we use the Chamfer distance to calculate MMD and COV.}
        \label{tab:5}
        \centering
        \setlength{\tabcolsep}{2pt}
        \renewcommand{\arraystretch}{1.3}
        \resizebox{1\linewidth}{!}{
        \begin{tabular}{lcccccc}
        \hline
        \multirow{2}{*}{Method} & \multicolumn{3}{c}{Chair}                            & \multicolumn{3}{c}{Airplane}                         \\
                                & FID $\downarrow$ & MMD $\downarrow$ & COV $\uparrow$ & FID $\downarrow$ & MMD $\downarrow$ & COV $\uparrow$ \\ \hline
        with PASD                 & 27.73            &          1.00   &        61.5    & 35.94            &      0.18        &     82.3       \\
        without PASD              & 30.09            &          1.03    &       61.0     & 38.84            &     0.19         &   81.8        \\ \hline
        \end{tabular}
        }
\end{table}

\textbf{Importance of the latent 3D diffusion}. 
Different from previous methods, this paper proposes the latent 3D diffusion that transfers the diffusion process into latent space, increasing the generative resolution significantly and preserving the high-quality generation.
We further apply the diffusion process to the voxel field directly to observe the performance change. Due to the limitation of computational cost, we set the resolution of voxel fields to 32$\times$32$\times$32. Tab.~\ref{tab:4} reports the performance comparison. Without the latent diffusion, the FID drops a lot, and the geometry metric of MMD and COV drop 0.40 and 6.4 on Chairs, which shows the importance of latent diffusion. Moreover, we visualize the qualitative results of the two different settings in Fig.~\ref{fig4-1}, and the results reveal that the latent 3D diffusion brings high-quality geometry and rendering details.

\begin{figure}
    \centering
    \includegraphics[width=\linewidth]{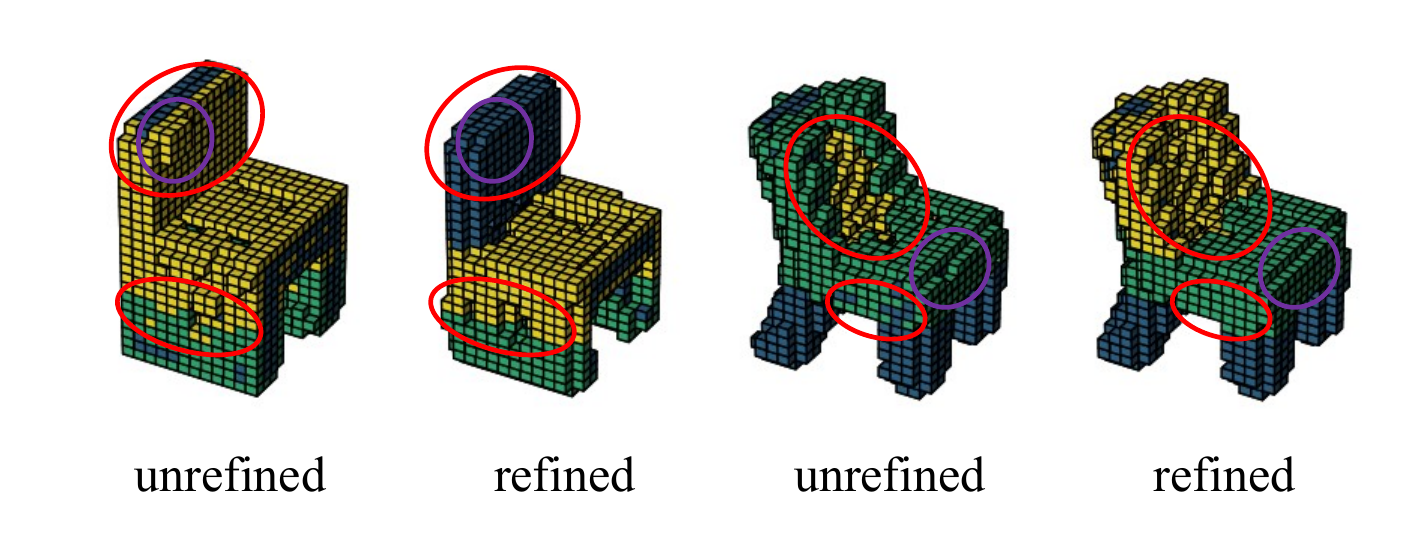}
    \caption{Comparisons between the voxel fields generated by the diffusion model and refined by the part-aware shape decoder.}
    \label{fig:refined}
\end{figure}

\textbf{Importance of the part-aware shape decoder}. The part-aware shape decoder incorporates the part code into the neural voxel field, generating the structural shape with accurate rendering results. To demonstrate its importance, we replace it with two independent MLPs, rendering the color and part maps from the voxel field directly. Tab.~\ref{tab:5} shows that the part-aware shape decoder can bring more improvements in rendering quality (FID) rather than geometry quality (MMD and COV). Since the part-aware shape decoder helps distinguish different local regions instead of the whole object geometry, the MMD and COV metrics improve marginally. As depicted in Fig.~\ref{fig4-2}, the part-aware shape decoder helps the model learn the accurate structural information, which allows for consistent rendering results of the local region. We also plot the visualization of the generated fields in Fig.~\ref{fig:refined}, including the generated fields by the diffusion model and the refined fields by the part-aware shape decoder. The part-aware shape decoder utilizes the cross- and self-attention to build relationships between different voxels, letting the model learn more discriminative features in each local part region and improving the part-aware generation. The red circle in Fig.~\ref{fig:refined} shows the improvements of local details in different parts and the purple circle shows that the refined fields are smoother.

\begin{table*}
    \caption{Ablation on different training strategy.
      }
  \label{tab:separate}
    \setlength{\tabcolsep}{3pt}
  \renewcommand{\arraystretch}{1.2}
  \centering
  \resizebox{1\textwidth}{!}{
    \begin{tabular}{lcccccccccccccccc}
    \hline
    \multirow{2}{*}{Method} & \multicolumn{2}{c}{Chair}       & \multicolumn{2}{c}{Table}       & \multicolumn{2}{c}{Airplane}    & \multicolumn{2}{c}{Car}         & \multicolumn{2}{c}{Guitar}      & \multicolumn{2}{c}{Lamp}        & \multicolumn{2}{c}{Laptop}      & \multicolumn{2}{c}{Pistol}      \\
                            & MMD$\downarrow$ & COV$\uparrow$ & MMD$\downarrow$ & COV$\uparrow$ & MMD$\downarrow$ & COV$\uparrow$ & MMD$\downarrow$ & COV$\uparrow$ & MMD$\downarrow$ & COV$\uparrow$ & MMD$\downarrow$ & COV$\uparrow$ & MMD$\downarrow$ & COV$\uparrow$ & MMD$\downarrow$ & COV$\uparrow$ \\ \hline
    Separate training                    & 1.24       & 58.2     & 1.45       & 56.9     & {0.31}       &  79.9    & {0.70}       &  33.6        & {0.25}        &  40.5      &   1.70       &     56.8    &  {0.89}        &   {35.3}     & {0.48}         &  {45.5}       \\
    Joint training                   & \textbf{1.00}       & \textbf{61.5}     & \textbf{1.13}       & \textbf{59.9}     & \textbf{0.18}       &  \textbf{82.3}    & \textbf{0.46}       &  \textbf{42.1}        & \textbf{0.14}        &  \textbf{45.1}      &   \textbf{1.48}       &     \textbf{59.7}    &  \textbf{0.79}        &   \textbf{38.8}     & \textbf{0.31}         &  \textbf{50.5}       \\ \hline
    \end{tabular}
    }
\end{table*}
\textbf{Importance of the gradient skip}. The gradient skip technique aims to reduce redundant gradient calculations and computational costs by jointly learning the latent 3D diffusion model and the part-aware shape decoder. In Fig.~\ref{fig-skip}, we compare our method with the separate training strategy, in which we train the diffusion model first and then the part-aware shape decoder. It is evident from the comparison that the separate training strategy lacking gradient skipping leads to the learning of a biased neural voxel field, consequently yielding inferior rendered images. Tab.~\ref{tab:separate} shows that the separate training strategy gets worse results in all categories.

\begin{figure*}[tbh]
    \centering
    \includegraphics[width=1\textwidth]{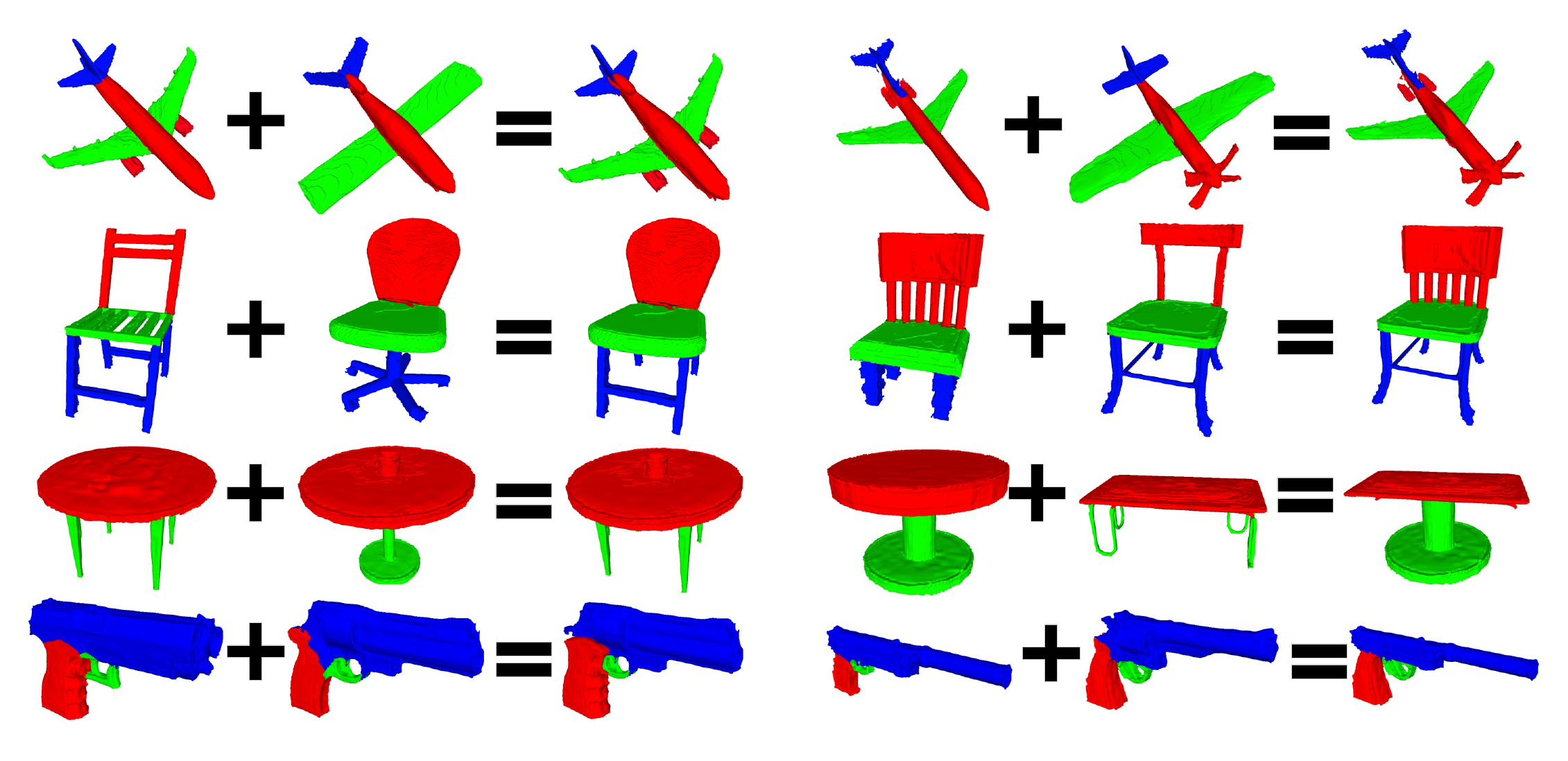}
    \caption{Visual results of shape mixing. Given two part-ware shapes, we can directly compose the different parts to create new reasonable shapes.}
    \label{fig8}
\end{figure*}

\begin{figure}[tb]
    \centering
    \includegraphics[width=\linewidth]{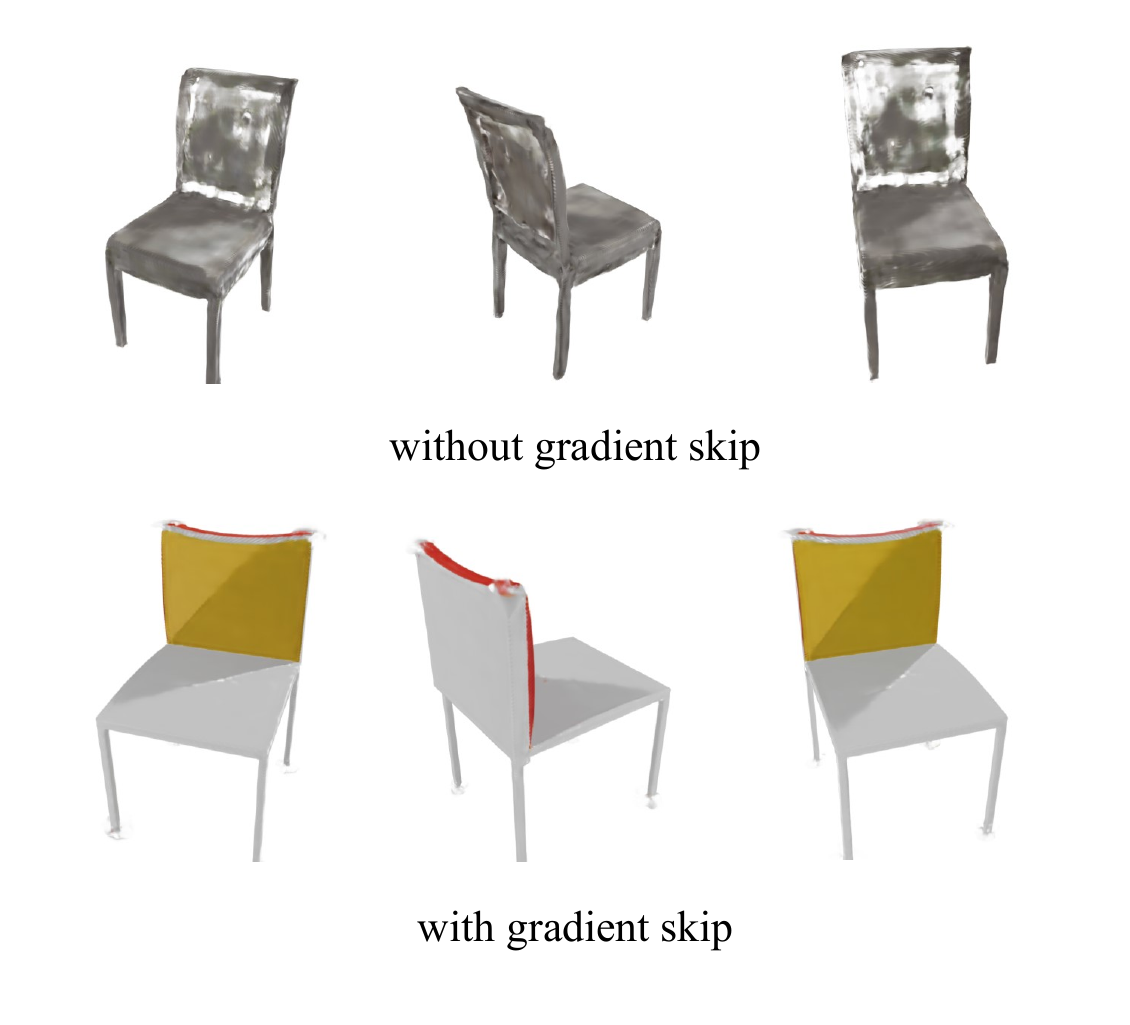}
    \caption{Visual improvements of the gradient skip.}
    \label{fig-skip}
\end{figure}

\textbf{Importance of 3D UNet}. We employ a 3D UNet as the network architecture to conduct the denoising process, and we have analyzed that the denoising network can be replaced by a 2D UNet possibly. However, we found that the 2D UNet failed to generate meaningful results as shown in Fig.~\ref{fig2}. Compared to 3D UNet, the training loss curve of 2D UNet is not converged, culminating in a noisy generation. This observation indicates that it is necessary to utilize the 3D convolutional operator to extract features for the 3D voxel fields.


\textbf{Effect of different diffusion models and sampling steps}. Firstly, we utilize typical diffusion models DDPM \cite{ddpm} and DDIM \cite{ddim} as our base diffusion model. As shown in Tab.~\ref{tab:6}, by leveraging rapid sampling from the decoupled diffusion model, our method only takes 10 steps to generate similar quality to DDIM with 30 steps, showing superior efficiency in generating high-quality shapes. 
Tab.~\ref{tab:6} presents a comparison of performance based on various sampling steps for chair generation. With an increase in the sampling step, performance shows a gradual improvement. Further increasing the sampling steps results in only a marginal improvement, suggesting that the generation quality is nearing saturation. 

\begin{table*}
    \caption{Ablation on different diffusion methods and sampling steps. Note that we report MMD / COV metrics on the Chair category.
      }
  \label{tab:6}
    \setlength{\tabcolsep}{5pt}
  \renewcommand{\arraystretch}{1.2}
  \centering
  \resizebox{0.7\textwidth}{!}{
    \begin{tabular}{lcccccccccccccccc}
    \hline
    Step & 10    & 20    & 30    & 40    & 50    \\ \hline
    DDPM  & 305.06 / 21.3 & 163.52 / 25.3 & 105.59 / 33.9 & 49.88 / 39.1 & 38.51/ 43.5 \\
    DDIM  & 1.32 / 54.1 & 1.19 / 58.5 & 1.10 / 59.4 & 1.03 / 61.2 & 1.00 / 61.6 \\
    Ours  & 1.10 / 59.0 & 1.05 / 60.6 & 1.00 / 61.5 & 1.00 / 61.6 & 0.99 / 61.7 \\ \hline
    \end{tabular}
    }
\end{table*}

\subsection{Shape Interpolation}
Following the 2D diffusion model \cite{ddim}, we can utilize two different voxel fields to generate intermediate shapes with spherical linear interpolation \cite{interpolation}. As shown in Fig.~\ref{fig5}, our model can achieve relatively smooth interpolation between the two shapes. In contrast, the model without the part-aware shape decoder generates unreasonable contents, which further indicates the effectiveness of the part-aware shape decoder. With the part-aware shape decoder, the generated intermediate shapes consistently maintain a precise structure, suggesting that the decoder effectively learns to decompose the neural voxel fields with the guidance of the part code.

\subsection{Single Image to Shape}
Single image to shape can be seen as a conditional generation task in which we use an image as conditional input and generate a shape related to the image. In practice, we extract the multi-scale image features via the Swin backbone network, and then concatenate the obtained features with the multi-scale features of the denoising UNet. As shown in Fig.~\ref{fig7}, given a reference image as a condition, our model can generate reasonable shapes with consistent geometry and accurate structure.


\subsection{Text-guided Shape Generation}
Text-guided shape generation is a form of conditioned generative method, where the text is embedded into a feature to control the generation process. We perform text-guided shape generation on Chairs and Tables. As shown in Fig.~\ref{fig:text}, our model is capable of generating reasonable shapes based on the provided texts. In fact, the textual condition provides a relatively coarse level of control for generation, resulting in better geometric accuracy compared to texture detail.


\subsection{Shape Mixing}
Due to the part-aware nature of the generated shapes, it becomes feasible to combine parts from different shapes to create new designs. Notably, these shapes conform to a shared coordinate system, enabling the direct integration of disparate parts to create a cohesive and compact shape. Fig.~\ref{fig8} illustrates the visual results of shape mixing on four different object categories. Our approach excels in producing clear part shapes, facilitating the composition of diverse shapes.




\section{Conclusion}
In this paper, we introduce a method for generating part-aware neural voxel fields using latent 3D diffusion models. The proposed method consists of a latent 3D diffusion model and a part-aware shape decoder. On the one hand, the latent 3D diffusion model enhances the high resolution of the neural voxel fields, thereby improving the geometric details and the quality of rendered images. We incorporate part codes into the latent 3D diffusion model, allowing the model to learn part-aware information in field generation.
On the other hand, the part-aware shape decoder integrates part codes into voxel fields to generate part-aware shapes with high-quality rendered images.
We evaluate the effectiveness of our method on four classes by comparing it against state-of-the-art diffusion-based approaches, demonstrating its efficacy in generating precise part-aware shapes.

\section{Acknowledgment}
This work is supported in part by the NSFC (62325211, 62132021, 62372457), the Major Program of Xiangjiang Laboratory (23XJ01009), Young Elite Scientists Sponsorship Program by CAST (2023QNRC001), the Natural Science Foundation of Hunan Province of China (2022RC1104) and the NUDT Research Grants (ZK22-52).


\bibliographystyle{IEEEtran} 
\bibliography{ref.bib} 










\newpage

\vspace{11pt}

\begin{IEEEbiography}[{\includegraphics[width=1in,height=1.25in,clip,keepaspectratio]{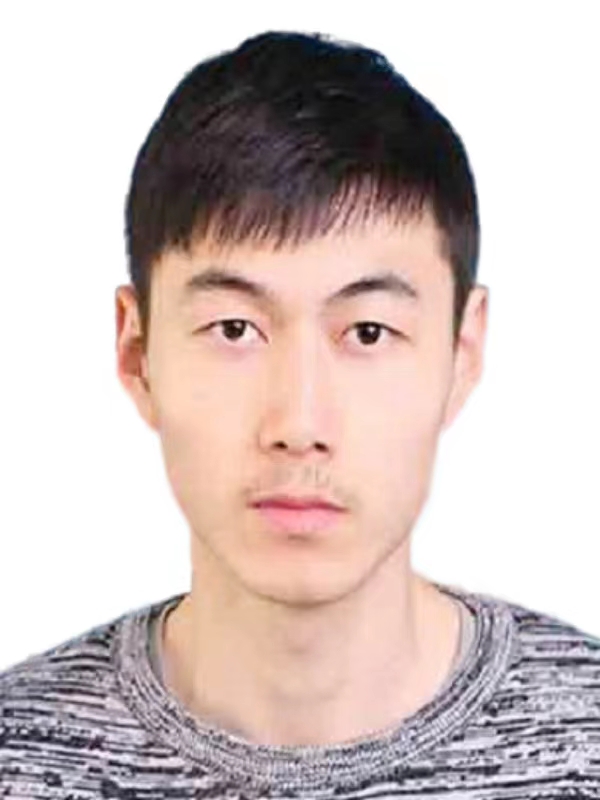}}]{Yuhang Huang}
received the bachelor’s degree from the University of Shanghai for Science and Technology, Shanghai, China, in 2019, and the master’s degree from Shanghai University, Shanghai, China, in 2022. He is currently pursuing the Ph.D. degree at the National University of Defense Technology, Changsha, China.

His research interests include computer vision, graphics, and generative models.
\end{IEEEbiography}

\vspace{11pt}
\begin{IEEEbiography}[{\includegraphics[width=1in,height=1.25in,clip,keepaspectratio]{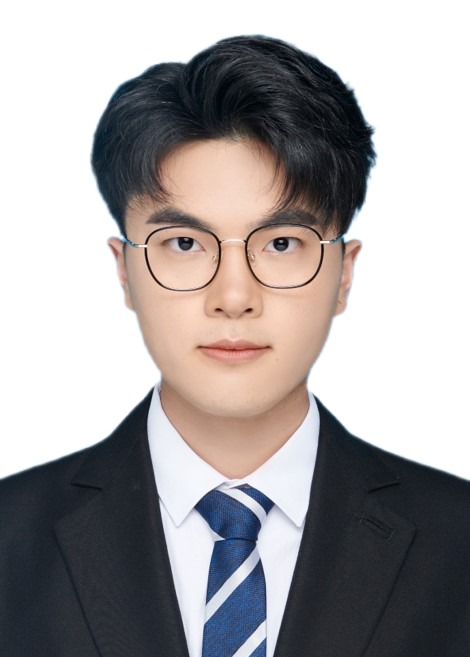}}]{Shilong Zou} received the bachelor's degree from the University of Dalian Maritime University, Dalian, 2023. He is currently pursuing the master's degree at the National University of Defense Technology, Changsha, China. 

His research interests include computer vision and generative models.
\end{IEEEbiography}

\vspace{11pt}
\begin{IEEEbiography}[{\includegraphics[width=1in,height=1.25in,clip,keepaspectratio]{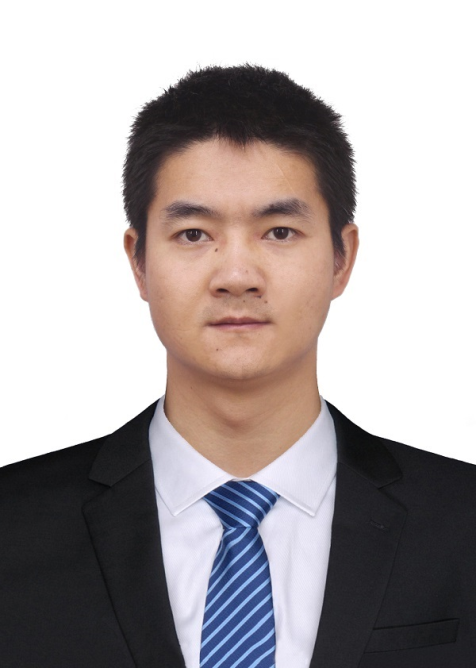}}]{Xinwang Liu}
received his PhD degree from National University of Defense Technology (NUDT), China. He is now Professor of School of Computer, NUDT. His current research interests include kernel learning and unsupervised feature learning. Dr. Liu has published 60+ peer-reviewed papers, including
those in highly regarded journals and conferences such as IEEE T-PAMI, IEEE T-KDE, IEEE T-IP, IEEE T-NNLS, IEEE T-MM, IEEE T-IFS, ICML, NeurIPS, ICCV, CVPR, AAAI, IJCAI, etc. He serves as the associated editor of Information Fusion Journal. More information can be found at \href{https://xinwangliu.github.io/}{https://xinwangliu.github.io/}.
\end{IEEEbiography}


\vspace{11pt}
\begin{IEEEbiography}[{\includegraphics[width=1in,height=1.25in,clip,keepaspectratio]{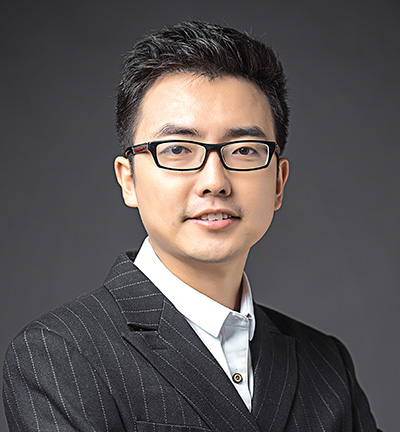}}]{Kai Xu} (Senior Member, IEEE) received the Ph.D. degree in computer science from the National University of Defense Technology (NUDT), Changsha,
China, in 2011. From 2008 to 2010, he worked as a Visiting Ph.D. degree with the GrUVi Laboratory, Simon Fraser University, Burnaby, BC, Canada. He is currently a Professor with the School of Computer Science, NUDT. He is also an Adjunct Professor with Simon Fraser University. His current research interests include data-driven shape analysis and modeling, and 3-D vision and robot perception and navigation.
\end{IEEEbiography}

\vfill

\end{document}